\documentclass[sigconf]{acmart}

\AtBeginDocument{%
  }

\copyrightyear{2022}
\acmYear{2022}
\setcopyright{acmcopyright}
\acmConference[CIKM '22] {Proceedings of the 31st ACM International Conference on Information and Knowledge Management}{October 17--21, 2022}{Atlanta, GA, USA.}
\acmBooktitle{Proceedings of the 31st ACM International Conference on Information and Knowledge Management (CIKM '22), October 17--21, 2022, Atlanta, GA, USA}
\acmPrice{15.00}
\acmISBN{978-1-4503-9236-5/22/10}

\usepackage[capitalize,noabbrev]{cleveref}

\theoremstyle{plain}

\theoremstyle{definition}

\theoremstyle{remark}

\usepackage{amsmath}
\usepackage{float}
\usepackage{amsthm}
\usepackage{subfigure}
\usepackage{pgfplots}
\usepackage{mathtools} 
\usepackage{algorithmic}
\usepackage{algorithm}
\usepackage{xcolor}
\usepackage{booktabs} 
\usepackage{url}
\usepackage{arydshln}
\renewcommand{\t}[1]{\tiny{#1}}

\usepackage{array}
\newcolumntype{P}[1]{>{\centering\arraybackslash}p{#1}}

\def\y{{\mathbf y}}

\def\X{{\mathbf X}}

\def\A{{\mathbf A}}
\def\L{{\mathbf L}}
\def\S{{\mathbf S}}
\def\D{{\mathbf D}}

\def\R{{\mathbb R}}

\begin{document}

\title{Higher-order Clustering and Pooling for Graph Neural Networks}

\author{Alexandre Duval}
\affiliation{%
  \institution{Université Paris-Saclay, CentraleSupélec, Inria}
  \country{Gif-sur-Yvette, France}
}
\email{alexandre.duval@centralesupelec.fr}

\author{Fragkiskos D. Malliaros}
\affiliation{%
  \institution{Université Paris-Saclay, CentraleSupélec, Inria}
  \country{Gif-sur-Yvette, France}
}
\email{fragkiskos.malliaros@centralesupelec.fr}

\begin{abstract}
Graph Neural Networks achieve state-of-the-art performance on a plethora of graph classification tasks, especially due to pooling operators, which aggregate learned node embeddings hierarchically into a final graph representation. However, they are not only questioned by recent work showing on par performance with random pooling, but also ignore completely higher-order connectivity patterns.
To tackle this issue, we propose \textsc{HoscPool}, a clustering-based graph pooling operator that captures higher-order information hierarchically, leading to richer graph representations. In fact, we learn a probabilistic cluster assignment matrix end-to-end by minimising relaxed formulations of motif spectral clustering in our objective function, and we then extend it to a pooling operator. We evaluate \textsc{HoscPool} on graph classification tasks and its clustering component on graphs with ground-truth community structure, achieving best performance. Lastly, we provide a deep empirical analysis of pooling operators' inner functioning. The code is available
\href{https://github.com/AlexDuvalinho/HoscPool}{here}.
\end{abstract}



\settopmatter{printacmref=False}

\maketitle

\section{Introduction}
\label{sec:intro}
Graph Neural Networks are powerful tools for graph datasets due to their message passing scheme, where they propagate node features along the edges of the graph to compute meaningful node representations \cite{hamilton2017inductive, kipf2016semi}. They achieve state-of-the-art performance on a variety of tasks including clustering, link prediction, node and graph classification \cite{zhou2020graph}. For the latter, since the goal is to predict the label of the entire graph, standard approaches pool together all nodes' embeddings to create a single graph representation, usually via a simple sum or average operation \cite{atwood2016diffusion}. This \textit{global pooling} discards completely graph structure when computing its final representation, failing to capture the topology of many real-world networks and thus preventing researchers to build effective GNNs. 

More desirable alternatives emerged to solve this limitation. They progressively coarsen the graph between message passing layers, for instance by regrouping highly connected nodes (i.e. clusters) together into supernodes with adapted adjacency / feature vectors. This allows to better capture the graph hierarchical structure compared to global pooling, without loosing relevant information if the coarsening is accurately done. While the first clustering-based pooling algorithms were deterministic \cite{defferrard2016convolutional, fey2018splinecnn} -- because of their high computational complexity, their transductive nature and their incapacity to leverage node features -- they were replaced by trainable \textit{end-to-end clustering} approaches such as \textsc{StructPool} \cite{yuan2020structpool} or \textsc{DiffPool} \cite{ying2018hierarchical}. Such methods solve the above limitations, often by learning a cluster assignment matrix along with GNN parameters thanks to a specific loss function, e.g. a link prediction score.

Despite presenting many advantages, such methods pool nodes together based on a simple functions or metrics which often lack strong supporting theoretical foundations. Besides, they reduce the graph uniquely based on first-order information. And in many cases, graph datasets may not present any edge-based connectivity structure, leading to insignificant graph coarsening steps, while they may have clear community structure with respect to more complex (domain-specific) motifs \cite{leskovec2009community}. Overall, this limits the expressiveness of the hierarchical information captured, and therefore of the classification performance. On top of that, existing pooling operators were surprisingly shown to perform on par with random pooling for many graph classification tasks, raising major concerns \cite{mesquita2020rethinking} and finding limited justifications. This discovery appears rather counter-intuitive as we logically expect the graph coarsening step, that is, the way to pool nodes together, to increase significantly the graph hierarchical information captured in its final representation. 


\begin{figure*}
    \centering
    \includegraphics[width=\textwidth]{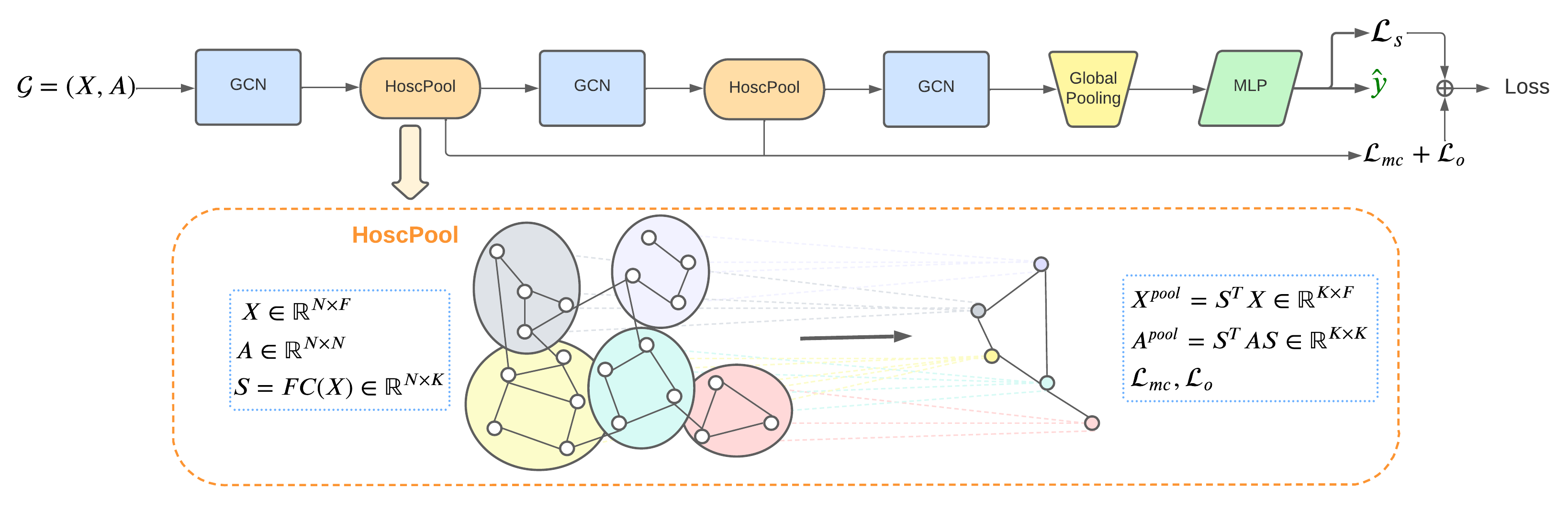}
    \caption{A graph classification pipeline with \textsc{HoscPool} hierarchical pooling to reduce graph $\mathcal{G}$ to $\mathcal{G}_{pool}=(\X_{pool}, \A_{pool})$ via a cluster assignment matrix $\S$ learned end-to-end from a motif spectral clustering inspired loss function $\mathcal{L}_{mc} + \mathcal{L}_{o}$.}
    \label{fig:hoscpool}
\end{figure*}

Combining these two facts, we propose \textsc{HoscPool}, a new end-to-end higher-order pooling operator grounded on probabilistic motif spectral clustering to capture a more advanced type of communities thanks to the incorporation of higher-order connectivity patterns. The latter has shown to be very successful for a wide range of applications \cite{lee2018higher} but has not yet been applied to graph classification, while it could greatly benefit from it. 
Specifically, we hierarchically coarsen the input graph using a cluster assignment matrix $\S$ learned by defining a well-motivated objective function, which includes continuous relaxations of motif conductance and thus combines various types of connectivity patterns for greater expressiveness. Since the process is fully differentiable, we can stack several such pooling layers, intertwined by message passing layers, to capture graph hierarchical information. We jointly optimise our unsupervised loss with any task-specific supervised loss function to allow truly end-to-end graph classification. Finally, we evaluate the performance of \textsc{HoscPool} on a plethora of graph datasets, and the reliability of its clustering algorithm on a variety of graphs endowed with ground-truth community structure. During this experiment phase, we proceed to a deep analysis aimed to understand why existing pooling methods fail to truly outperform random baselines and attempt to provide explications. This is another important contribution, which we hope will help future works.

\section{Related Work}
\label{sec:rw}
\textbf{Graph pooling}. Leaving aside global pooling \cite{atwood2016diffusion, simonovsky2017dynamic, xu2018powerful}, we distinguish between two main types of hierarchical approaches. Node drop methods  \cite{lee2019self, gao2019graph, zhang2018end, pang2021graph, baek2021accurate, 9534320, xu2022multistructure} use a learnable scoring function based on message passing representations to assess all nodes and drop the ones with lowest score. The drawback is that we loose information during pooling by dropping completely certain nodes. On the other hand, clustering approaches cast the pooling problem as a clustering one \cite{ma2019graph, diehl2019edge, luzhnica2019clique, wang2020haar, yuan2020structpool, ranjan2020asap, liu2021hierarchical}. For instance, \textsc{StructPool} \cite{yuan2020structpool}  utilizes conditional random fields to learn the cluster assignment matrix; \textsc{HaarPool} \cite{wang2020haar} uses the compressive Haar transform; \textsc{EdgePool} \cite{diehl2019edge} gradually merges nodes by contracting high-scoring edges. Of particular interest here are two very popular end-to-end clustering methods, namely \textsc{DiffPool} \cite{ying2018hierarchical} and \textsc{MinCutPool} \cite{bianchi2020spectral}, because of their original and efficient underlying idea. While \textsc{DiffPool} utilises a link prediction objective along with an entropy regularization to learn the cluster assignment matrix, \textsc{MinCutPool} leverages an min-cut score objective along with an orthogonality term. Although there are more pooling operators, we wish to improve this line of method, that we think is promising and perfectible. In addition to solving existing limitations, we want to introduce the notion of higher-order to pooling for graph classification, which is unexplored yet. \\


\textbf{Higher-order} connectivity patterns (i.e. motifs -- small network subgraphs like triangles \includegraphics[scale=0.055]{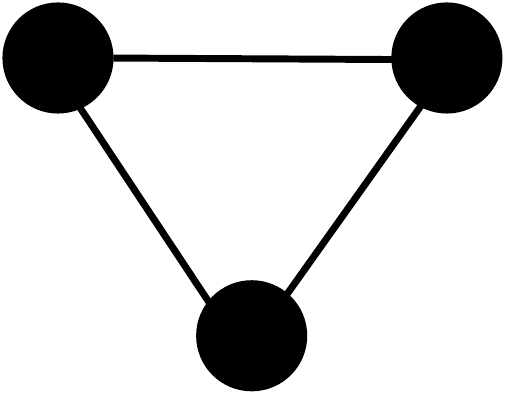}).), are known to be the fundamental building blocks of complex networks \cite{milo2002network, carranza2020higher}. They are essential for modelling and understanding the organization of various types of networks. For instance, they play an essential role in the characterisation of social, biological or molecules networks \cite{morris2019weisfeiler}. \cite{eswaran2020higher} showed that vertices participating in the same higher-order structure often share the same label, spreading its adoption to node classification tasks \cite{lee2018higher, li2021higher}. Going further, several recent research papers have clearly demonstrated the benefits of leveraging higher-order structure for link prediction \cite{abuoda2019link, sharma2021higher}, explanation generation \cite{schnake2020higher, perotti2022graphshap}, ranking \cite{rossi2019higher}, clustering \cite{klymko2014using, hu2021hiscf}. Regarding the latter, \cite{tsourakakis2017scalable, benson2016higher} argue that domain-specific motifs are a better signature of the community structure than simple edges. Their intuition is that motifs allow us to focus on particular network substructures that are important for networks of a given domain. As a result, they generalized the notion of conductance to triangle conductance (Section \ref{sec:background}), which was found highly beneficial by \cite{carranza2020higher, sotiropoulos2021triangle}. 



\section{Preliminary Knowledge}
\label{sec:background}
$\mathcal{G}=(\mathcal{V},\mathcal{E})$ is a graph with vertex set $\mathcal{V}$ and edge set $\mathcal{E}$, characterised by its adjacency matrix $\A \in \mathbb{R}^{N\times N}$ and node feature matrix $\X \in \mathbb{R}^{N\times F}$. $\D = \text{diag}(\A \boldsymbol{1}_N)$ is the degree matrix and $\L = \D - \A $ the Laplacian matrix of $\mathcal{G}$. $\tilde{\A} = \D^{-\frac{1}{2}}\A\D^{-\frac{1}{2}} \in \R^{N \times N}$ is the symmetrically normalised adjacency matrix with corresponding $\tilde{\D}$, $\tilde{\L}$. 



\subsection{Graph Cut and Normalised Cut}

Clustering involves partitioning the vertices of a graph into $K$ disjoint subsets with more intra-connections than inter-connections \cite{von2007tutorial}. One of the most common and effective way to do it \cite{schaeffer2007graph} is to solve the Normalised Cut problem \cite{shi2000normalized}: 
\begin{equation}
    \label{Ncut}
    \min_{\mathcal{S}_1, \ldots, \mathcal{S}_K}  \sum_{k=1}^K \frac{\texttt{cut}(\mathcal{S}_k, \bar{\mathcal{S}_k})}{\texttt{vol}(\mathcal{S}_k)},
\end{equation}
\noindent where $\bar{\mathcal{S}_k}=\mathcal{V} \setminus \mathcal{S}_k$, $\texttt{cut}(\mathcal{S}_k, \bar{\mathcal{S}_k}) = \sum_{i \in \mathcal{S}_K, j \in \bar{\mathcal{S}}_K} A_{ij}$, and $\texttt{vol}(\mathcal{S}_k)= \sum_{i \in \mathcal{S}_k, j \in \mathcal{V}} A_{ij}$. Unlike the simple min-cut objective, 
(\ref{Ncut}) scales each term by the cluster volume, thus enforcing clusters to be ``reasonably large'' and avoiding degenerate solutions where most nodes are assigned to a single cluster. Although minimising (\ref{Ncut}) is NP-hard \cite{wagner1993between}, there are approximation algorithms with theoretical guarantees \cite{chung2007four} for finding clusters with small conductance, such as Spectral Clustering (SC), which proposes clusters determined based on the eigen-decomposition of the Laplacian matrix. A refresher on SC is provided in \cite{von2007tutorial}.


\subsection{Motif conductance}

While the Normalised Cut builds on first-order connectivity patterns (i.e. edges), \cite{benson2016higher, tsourakakis2017scalable} propose to cluster a network based on specific higher-order substructures. Formally, for graph $\mathcal{G}$, motif $M$ made of $|M|$ nodes, and $\mathcal{M}= \{ \mathbf{v} \in \mathcal{V}^{|M|} | \mathbf{v}=M \} $ the set of all instances of $M$ in $\mathcal{G}$, they propose to search for the partition $\mathcal{S}_1, \ldots, \mathcal{S}_K$ minimising motif conductance:
\begin{equation}
    \label{motif-conductance}
    \min_{\mathcal{S}_1, \ldots, \mathcal{S}_K} \sum_{k=1}^K \dfrac{\texttt{cut}^{(\mathcal{G})}_M(\mathcal{S}_k,\bar{\mathcal{S}_k})} {\texttt{vol}^{(\mathcal{G})}_M(\mathcal{S}_k)},   
\end{equation}
where $\texttt{cut}^{(\mathcal{G})}_M(\mathcal{S}_k,\bar{\mathcal{S}_k}) = \sum_{\mathbf{v} \in \mathcal{M}} \mathbf{1}(\exists i,j \in \mathbf{v} |i\in \mathcal{S}_k, j\in \bar{\mathcal{S}_k})$, i.e. the number of instances $\mathbf{v}$ of $M$ with at least one node in $\mathcal{S}_k$ and at least one node in $\bar{\mathcal{S}_k}$; and  $\texttt{vol}^{(\mathcal{G})}_M(\mathcal{S}_k) = \sum_{\mathbf{v} \in \mathcal{M}} \sum_{i \in \mathbf{v}} \mathbf{1}(i \in \mathcal{S}_k)$, i.e. the number of motif instance endpoints in $\mathcal{S}_k$. 

\section{Proposed Method}
\label{sec:method}
The objective of this paper is to design a differentiable cluster assignment matrix $\S$ that learns to find relevant clusters based on higher-order connectivity patterns, in an end-to-end manner within any GNN architecture. To achieve this, we formulate a continuous relaxation of motif spectral clustering and embed the derived formulation into the model objective function to enforce its learning.


\subsection{Probabilistic motif spectral clustering}
\label{subsec:motif-spectral-clustering}

Before exploring how we can rewrite the motif conductance optimisation problem (\ref{motif-conductance}) in a solvable way, we introduce the motif adjacency matrix $\A_M$, where each entry $(A_M)_{ij}$ represents the number of motifs in which both node $i$ and node $j$ participate. Its diagonal has zero values. Formally, $(A_M)_{ij} = \sum_{\mathbf{v} \in \mathcal{M}} \mathbf{1}(i,j \in \mathbf{v}, i \neq j)$. $\mathcal{G}_M$ is the graph induced by $\A_M$. $(D_M)_{ii} = \sum_{j=1}^N (A_M)_{ij}$ and $\L_M$ are the motif degree and motif Laplacian matrices.

For now, we focus on triangle motifs ($M = K_3$), and extend to more complex motifs in Section \ref{subsec:end-to-end}. From \cite{benson2016higher}, we have:
\begin{align*}
    &\texttt{cut}_M^{(\mathcal{G})}(\mathcal{S}_k, \bar{\mathcal{S}_k}) = \frac{1}{2} \sum_{i \in \mathcal{S}_k} \sum_{j \in \bar{S}_k} (A_M)_{ij} \\
    &\texttt{vol}_M^{(\mathcal{G})}(\mathcal{S}_k) = \frac{1}{2} \sum_{i \in \mathcal{S}_k} \sum_{j \in \mathcal{V}} (A_M)_{ij},
\end{align*}
which enables us to rewrite (\ref{motif-conductance}) as:
%
\begin{align}
\label{eq:max-motif-conductance}
    & \min_{\mathcal{S}_1, \ldots, \mathcal{S}_K} \sum_{k=1}^K \frac{ \sum_{i \in \mathcal{S}_k, j \in \bar{\mathcal{S}_k}} (A_M)_{ij} }{ \sum_{i \in \mathcal{S}_k, j \in \mathcal{V}} (A_M)_{ij} } \nonumber \\
    \equiv & \max_{\mathcal{S}_1, \ldots, \mathcal{S}_K} \sum_{k=1}^K \frac{ \sum_{i,j \in \mathcal{S}_k} (A_M)_{ij}}{\sum_{i \in \mathcal{S}_k, j \in \mathcal{V}} (A_M)_{ij}}, 
\end{align}
where the last equivalence follows from 
\begin{align*}
    \sum_{i,j \in \mathcal{S}_k} (A_M)_{ij} +  \sum_{i \in \mathcal{S}_k, j \in \bar{\mathcal{S}_k}} (A_M)_{ij} = \sum_{i \in \mathcal{S}_k, j \in \mathcal{V}} (A_M)_{ij}. 
\end{align*}

Instead of using partition sets, we define a discrete cluster assignment matrix $\S \in \{0,1\}^{N\times K}$ where $\S_{ij}=1$ if $v_i \in \mathcal{S}_j$ and $0$ otherwise. We denote by $\S_j=[S_{1j}, \ldots, S_{Nj}]^\top$ the $j^{th}$ column of $\S$, which indicates the nodes belonging to cluster $\mathcal{S}_j$. Using this, we transform (\ref{eq:max-motif-conductance}) into:
%
\begin{align}
    \label{probab-motif-condutance}
     &\max_{\S \in \{0,1\}^{N\times K}} \sum_{k=1}^K \frac{ \sum_{i,j \in \mathcal{V}} (A_M)_{ij} S_{ik} S_{jk}}{ \sum_{i,j \in \mathcal{V}} S_{ik} (A_M)_{ij} } \nonumber \\
    \equiv &\max_{\S \in \{0,1\}^{N\times K}} \sum_{k=1}^K \frac{ \S_k^\top \A_M  \S_k}{\S_k^\top \D_M  \S_k} \nonumber \\ 
    \equiv & \min_{\S \in \{0,1\}^{N\times K}} - \text{Tr} \bigg(\frac{ \S^\top \A_M \S}{\S^\top \D_M  \S}\bigg), 
\end{align}
where the division sign in the last line is an element-wise division on the diagonal of both matrices. By definition, $\S$ is subject to the constraint $\S \mathbf{1}_K = \mathbf{1}_N$, i.e. each node belongs exactly to 1 cluster.

This optimisation problem is NP-hard since $\S$ take discrete values. We thus relax it to a probabilistic framework, where $\S$ take continuous values in the range $[0,1]$, representing cluster membership probabilities, i.e. each entry $S_{ik}$ denotes the probability that node $i$ belongs to cluster $k$. Referring to \cite{von2007tutorial} and \cite{benson2016higher}, solving this continuous relaxation of motif spectral clustering approximates a closed form solution with theoretical guarantees, provided by the Cheeger inequality \cite{chung2007four}. Compared to the original hard assignment problem, this soft cluster assignment formulation is less likely to be trapped in local minima \cite{jin2005probabilistic}. It also allows to generalise easily to multi-class assignment, expresses uncertainty in clustering, and can be optimised within any GNN.


\subsection{End-to-end clustering framework}
\label{subsec:end-to-end}

In this section, we leverage this probabilistic approximation of motif conductance to learn our cluster assignment matrix $\S$ in a trainable manner. Our method addresses the limitations of (motif) spectral clustering: we cluster nodes based both on graph topology and node features; leverage higher-order connectivity patterns; avoid the expensive eigen-decomposition of the motif Laplacian; and allow to cluster out-of-sample graphs.

We compute the soft cluster assignment matrix $\S$ using one (or more) fully connected layer(s), mapping each node's representation $\X_{i*}$ to its probabilistic cluster assignment vector $\S_{i*}$. We apply a softmax activation function to enforce the constraint inherited from (\ref{probab-motif-condutance}): $S_{ij} \in [0,1]$ and $\S\boldsymbol{1}_K = \boldsymbol{1}_N$: 
\begin{equation}
    \label{eq:assigments}
    \begin{aligned}
        \S & = \text{FC}(\X; \boldsymbol{\Theta}).  
    \end{aligned}
\end{equation}
$\boldsymbol{\Theta}$ are trainable parameters, optimised by minimising the unsupervised loss function $\mathcal{L}_{mc}$, which approximates the relaxed formulation of the motif conductance problem (\ref{probab-motif-condutance}):
\begin{equation}
    \label{eq:loss-motif-conductance}
    \mathcal{L}_{mc}=  - \frac{1}{K} \cdot \text{Tr}\bigg(\frac{\S^\top\A_{M}\S}{\S^\top\D_M\S}\bigg).
\end{equation}
Referring to the spectral clustering formulation\footnote{The largest eigenvalue $\A_M\S=\lambda \D_M\S$ is 1 and the smallest 0; we are summing only the $k$ largest eigenvalues.}, $\mathcal{L}_{mc} \in [-1,0]$. It reaches $-1$ when $\mathcal{G}_M$ has $\geq K$ connected components (no motif endpoints are separated by clustering), and 0 when for each pair of nodes participating in the same motif (i.e. $(A_M)_{ij}>0$), the cluster assignments are orthogonal: $\langle \S_{i*}, \S_{j*} \rangle = 0$. $\mathcal{L}_{mc}$ is a non-convex function and its minimisation can lead to local minima, although our probabilistic membership formulation makes it less likely to happen w.r.t. hard membership \cite{jin2005probabilistic}. 

In fact, we allow the \textit{combination of several motifs} inside our objective function (\ref{eq:loss-motif-conductance}) via $\mathcal{L}_{mc}= \sum_j \alpha_j \mathcal{L}_{{mc}^j}$ where $\mathcal{L}_{{mc}^j}$ denotes the objective function with respect to a particular motif (e.g., edge \includegraphics[scale=0.055]{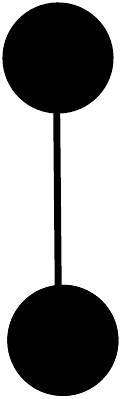}, triangle \includegraphics[scale=0.055]{data/3-triangle.pdf}, 4-nodes cycle \includegraphics[scale=0.055]{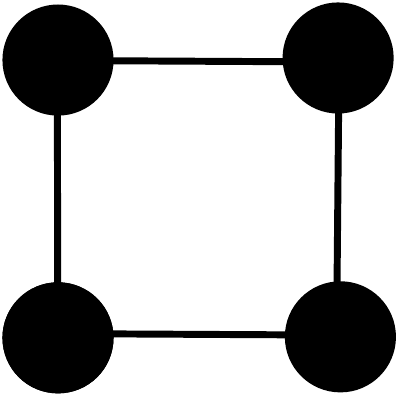}) and $\alpha_j$ is an importance factor. 
This also increases the power of our method, allowing us to find communities of nodes w.r.t. a hierarchy of higher-order substructures. As a result, the graph coarsening step will pool together more relevant groups of nodes, potentially capturing more relevant patterns in subsequent layers, ultimately producing richer graph representation. We implement it for edge and triangle motifs:
\begin{align}
    \mathcal{L}_{mc} = - \frac{\alpha_1}{K} \cdot \text{Tr}\bigg(\frac{\S^\top\A\S}{\S^\top\D\S}\bigg) - \frac{\alpha_2}{K} \cdot \text{Tr}\bigg(\frac{\S^\top\A_{M}\S}{\S^\top\D_{M}\S}\bigg).
    \label{eq:mixedpool}
\end{align}
We let $\alpha_1$, $\alpha_2$, to be dynamic functions of the epoch, subject to $\alpha_1 + \alpha_2=1$, allowing to first optimise higher-order motifs before moving on to smaller ones. It helps refine the level of granularity progressively and was found desirable empirically. This is the higher-order clustering formulation that we consider in the paper. 

In case we would like to enforce more rigorously the hard cluster assignment, characteristic of the original motif conductance formulation, we design an auxiliary loss function:
\begin{equation}
    \label{eq:ortho-loss}
    \mathcal{L}_o = \frac{1}{\sqrt{K}-1} \bigg( \sqrt{K} - \frac{1}{\sqrt{N}}\sum_{j=1}^K ||S_{*j}||_F\bigg),
\end{equation}
where $\| \cdot \|_F$ indicates the Frobenius norm. This orthogonality loss encourages more balanced and discrete clusters (i.e. a node assigned to a cluster with high probability, while to other clusters with a low one), discouraging further degenerate solutions. Although its effect overlaps with $\mathcal{L}_{mc}$, it often smoothes out the optimisation process and even improves slightly performance in complex tasks or networks, such as graph classification. In (\ref{eq:ortho-loss}), we rescale $\mathcal{L}_o$ to $[0,1]$, making it commensurable to $\mathcal{L}_{mc}$. As a result, the two terms can be safely summed and optimised together when specified. A parameter $\mu$ controls the strength of this regularisation. 

Similarly to other cluster-based pooling operators, our method relies on two assumptions. Firstly, nodes are identifiable via their features. Secondly, node features represent a good initialisation for computing cluster assignments. The latter is realistic due to the homophily property of many real-world networks \cite{mcpherson2001birds} as well as the smoothing effect of message passing layers \cite{chen2020measuring}, which render connected nodes more similar.


We conclude this section with a note for future work. An interesting research direction would be to extend this framework to 4-nodes motifs. Despite having managed to derive a theoretical formulation for the 4-nodes motif conductance problem in Appendix \ref{subsec: extension-4-nodes}, it becomes complex and would probably necessitate its own dedicated research, as it could be an promising extension. 


\subsection{Higher-order graph coarsening}
\label{subsec:pooling}

The methodology detailed in the previous sections is a general clustering technique that can be used for any clustering tasks on any graph dataset. In this paper, we utilise it to form a pooling operator, called \textsc{HoscPool}, which exploits the cluster assignment matrix $\S$ to generate a coarsened version of the graph (with fewer nodes and edges) that preserve critical information and embeds higher-order connectivity patterns. More precisely, it coarsens the existing graph by creating super-nodes from the derived clusters, with a new edge set and feature vector, depending on previous nodes belonging to this cluster. Mathematically,
\begin{align*}
\textsc{HoscPool}&: \mathcal{G}=(\X, \A) \rightarrow \mathcal{G}^{pool}=(\X^{pool}, \A^{pool}) \\
\A^{pool}&=\S^\top\A\S \text{~~ and ~~ } \X^{pool}=\S^\top {\X}.  
\end{align*}
Each entry $X_{i,j}^{pool}$ denotes feature $j$'s value for cluster $i$, calculated as a sum of feature $j$'s value for the nodes belonging to cluster $i$, weighted by the corresponding cluster assignment scores.  
$\A^{pool} \in \R^{K \times K}$ is a symmetric matrix where $A^{pool}_{i,j}$ can be viewed as the connection strength between cluster $i$ and cluster $j$.
Given our optimisation function, it will be a diagonal-dominant matrix, which will hamper the propagation across adjacent nodes. For this reason, we remove self-loops. We also symmetrically normalise the new adjacency matrix. Lastly, note that we use the original $\A$ and $\X$ for this graph coarsening step; their motif counterparts $\A_M$ and $\X_M$ are simply leveraged to compute the loss function. Our work thus differ clearly from diffusion methods and traditional GNN leveraging higher-order. 

Because our GNN-based implementation of motif spectral clustering is fully differentiable, we can stack several \textsc{HoscPool} layers, intertwined with message passing layers, to hierarchically coarsen the graph representation. In the end, a global pooling and some dense layers produce a graph prediction. The parameters of each \textsc{HoscPool} layer can be learned end-to-end by jointly optimizing:
\begin{equation}
    \mathcal{L} = \mathcal{L}_{mc} + \mu \mathcal{L}_{o} + \mathcal{L}_s, 
\end{equation}
where $\mathcal{L}_s$ denotes  any supervised loss for a particular downstream task (here the cross entropy loss). This way, we should be able to hierarchically capture relevant graph higher-order structure while learning GNN parameters so as to ultimately better classify the graphs within our dataset. 



\subsection{Comparison with relevant baselines}
\label{subsec:comparison}

\begin{figure}[!t]
    \centering
    \begin{minipage}[t]{\columnwidth}
        \centering
        \begin{tikzpicture}[scale=0.93]
        	\begin{axis}[
        	    title=\textsc{HoscPool},
        	    width=\columnwidth, 
        	    height=0.6*\axisdefaultheight,
        	    title style={at={(0.5,0.96)}},
        	    legend style={at={(0.6,0.4)},anchor=west},
        		ylabel near ticks,
        		ylabel=Loss value,
        		legend style={font=\small},
        		x tick label style={font=\small},
        		y tick label style={font=\small},
        		xlabel style={font=\small, at={(axis description cs:1.02, 0.16)}},
        		ylabel style={font=\small},
        		xmax=69,
        		xmin=0]
        		\hspace{-.2cm}
        	\addplot[color=blue, thick] coordinates {
                (1,	-0.143021584)
                (3,	-0.196151346)
                (5,	-0.338714421)
                (7,	-0.499774754)
                (9,	-0.639657021)
                (11, -0.730783761)
                (13, -0.777307272)
                (15,	-0.799121022)
                (17,	-0.818313301)
                (19,	-0.848052561)
                (21,	-0.8929618)
                (23,	-0.922325134)
                (25,	-0.931799531)
                (27,	-0.938346088)
                (29,	-0.94321233)
                (31,	-0.950235546)
                (33,	-0.953860641)
                (35,	-0.956063032)
                (37,	-0.967540085)
                (39,	-0.971409023)
                (41,	-0.975188732)
                (43,	-0.972756565)
                (45,	-0.973696828)
                (47,	-0.972769022)
                (49,	-0.97468406)
                (51,	-0.976153016)
                (53,	-0.977561593)
                (55,	-0.979072332)
                (57,	-0.977800548)
                (59,	-0.970999479)
                (61,	-0.972234488)
                (63,	-0.977481723)
                (65,	-0.976459384)
                (67,	-0.977197766)
                (69,	-0.978032827)
        	};
        	\addlegendentry{objective}
        	\addplot[color=red, thick, dashed] coordinates{
            	(2,	0.995914117)
                (4,	0.887462348)
                (6,	0.792716965)
                (8,	0.728382766)
                (10,	0.685405955)
                (13,	0.651383176)
                (15,	0.623745285)
                (17,	0.604108684)
                (19,	0.587963238)
                (21,	0.574848838)
                (24,	0.56273438)
                (26,	0.551557317)
                (28,	0.540053435)
                (30,	0.53043671)
                (32,	0.519517474)
                (35,	0.511612296)
                (37,	0.504185073)
                (39,	0.496467724)
                (41,	0.489894412)
                (43,	0.482332222)
                (46,	0.475239195)
                (48,	0.468538366)
                (50,	0.453497303)
                (52,	0.448354689)
                (54,	0.444111099)
                (57,	0.440898111)
                (59,	0.437053425)
                (61,	0.435165895)
                (63,	0.432630015)
                (65,	0.429878963)
                (68,	0.427380709)
                (70,	0.423801524)
                };
        	\addlegendentry{regularizer}
        	\end{axis}
        \end{tikzpicture}
    \end{minipage}
    \begin{minipage}[t]{\columnwidth}
        \centering
        \vspace{-0.3cm}
        \begin{tikzpicture}[scale=0.93]
        \centering
        	\begin{axis}[
        	    title=\textsc{MinCutPool},
        	    title style={at={(0.5,0.96)}},
        	    width=\columnwidth, 
        	    height=0.6*\axisdefaultheight,
        	    legend style={at={(0.6,0.4)},anchor=west},
        	    legend style={font=\small},
        	    x tick label style={font=\small},
        		y tick label style={font=\small},
        		xlabel style={font=\small, at={(axis description cs:0.5, 0.05)}},
        		y label style={at={(axis description cs:0,-.9)},anchor=north},
        		ylabel near ticks,
        		ylabel style={font=\small},
        		ylabel=Loss value,
        		xmax=69,
        		xmin=0]
        	\hspace{-.2cm}	
        	\addplot[color=blue, thick] coordinates {
            (2,	-0.999453187) 
            (4,	-0.999986291) 
            (6,	-0.999993563) 
            (8,	-0.999970257) 
            (10,	-0.999736845) 
            (12,	-0.998301148) 
            (14,	-0.99443388) 
            (16,	-0.989476979) 
            (18,	-0.983730316) 
            (20,	-0.979306281) 
            (22,	-0.976258457) 
            (24,	-0.97441709) 
            (26,	-0.973244905) 
            (28,	-0.971931338) 
            (30,	-0.970392525) 
            (32,	-0.968555987) 
            (34,	-0.966656923) 
            (36,	-0.965353966) 
            (38,	-0.963962078) 
            (40,	-0.962515235) 
            (42,	-0.961044014) 
            (44,	-0.959730506) 
            (46,	-0.959026515) 
            (48,	-0.958359718) 
            (50,	-0.957728505) 
            (52,	-0.957139015) 
            (54,	-0.956616223) 
            (56,	-0.956356287) 
            (58,	-0.956109107) 
            (60,	-0.955875099) 
            (62,	-0.955652952) 
            (64,	-0.95543915) 
            (66,	-0.955337167) 
            (68,	-0.95523715) 
            (70,	-0.955141246) 
            (72,	-0.955048203) 
            (74,	-0.954957843) 
            (76,	-0.954897702) 
            (78,	-0.954842448) 
            (80,	-0.954787493)
        	};
        	\addlegendentry{objective}
        	\addplot[color=red, thick, dashed] coordinates{
        	(2,	0.999817729)
            (4,	0.999990106)
            (6,	0.999987304)
            (8,	0.999893308)
            (10,	0.999008298)
            (12,	0.99343735)
            (14,	0.978235483)
            (16,	0.956757963)
            (18,	0.925737679)
            (20,	0.892678976)
            (22,	0.863847315)
            (24,	0.843548656)
            (26,	0.83113277)
            (28,	0.818062186)
            (30,	0.803445935)
            (32,	0.786384046)
            (34,	0.769296587)
            (36,	0.757973075)
            (38,	0.745912254)
            (40,	0.733216584)
            (42,	0.719962299)
            (44,	0.707649767)
            (46,	0.700749755)
            (48,	0.693909287)
            (50,	0.6871503)
            (52,	0.680477619)
            (54,	0.674212039)
            (56,	0.670965314)
            (58,	0.667778313)
            (60,	0.664649606)
            (62,	0.661572933)
            (64,	0.658544362)
            (66,	0.657053947)
            (68,	0.655586779)
            (70,	0.654141724)
            (72,	0.652715623)
            (74,	0.651306868)
            (76,	0.650364816)
            (78,	0.649482787)
            (80,0.648609459)
        	};
        	\addlegendentry{regularizer}
        	\end{axis}
        \end{tikzpicture}
    \end{minipage}
    \begin{minipage}[t]{\columnwidth}
        \centering
        \vspace{0cm}
        \begin{tikzpicture}[scale=0.93]
        \centering
        	\begin{axis}[
        	    title=\textsc{MinCutPool - degenerate solution},
        	    title style={at={(0.5,0.96)}},
        	    width=\columnwidth, 
        	    height=0.6*\axisdefaultheight,
        	    legend style={at={(0.6,0.4)},anchor=west},
        	    legend style={font=\small},
        	    x tick label style={font=\small},
        		y tick label style={font=\small},
        	    xlabel=Epochs,
        		xlabel style={font=\small, at={(axis description cs:0.5, 0.05)}},
        		y label style={at={(axis description cs:0,-.9)},anchor=north},
        		ylabel near ticks,
        		ylabel style={font=\small},
        		ylabel=Loss value,
        		xmax=69,
        		xmin=0]
        	\hspace{-.2cm}	
        	\addplot[color=blue, thick] coordinates {
            (2,	-0.999453187) 
            (4,	-0.999986291) 
            (6,	-0.999993563) 
            (8,	-0.999970257) 
            (10,	-0.999736845) 
            (12,	-0.998301148) 
            (14,	-0.99443388) 
            (16,	-0.99443388) 
            (18,	-0.99443388) 
            (20,	-0.99443388) 
            (22,	-0.99443388) 
            (24,	-0.99443388) 
            (26,	-0.983244905) 
            (28,	-0.983244905) 
            (30,	-0.983244905) 
            (32,	-0.983244905) 
            (34,	-0.983244905) 
            (36,	-0.983244905) 
            (38,	-0.983244905) 
            (40,	-0.983244905) 
            (42,	-0.983244905) 
            (44,	-0.983244905) 
            (46,	-0.983244905) 
            (48,	-0.983244905) 
            (50,	-0.983244905) 
            (52,	-0.983244905) 
            (54,	-0.983244905) 
            (56,	-0.983244905) 
            (58,	-0.983244905) 
            (60,	-0.983244905) 
            (62,	-0.983244905) 
            (64,	-0.983244905) 
            (66,	-0.983244905) 
            (68,	-0.983244905) 
            (70,	-0.983244905) 
            (72,	-0.983244905) 
            (74,	-0.983244905) 
            (76,	-0.983244905) 
            (78,	-0.983244905) 
            (80,	-0.983244905) 
        	};
        	\addlegendentry{objective}
        	\addplot[color=red, thick, dashed] coordinates{
        	(2,	0.999817729)
            (4,	0.999990106)
            (6,	0.999987304)
            (8,	0.999893308)
            (10,	0.999008298)
            (12,	0.999008298)
            (14,	0.999008297)
            (16,	0.999008293)
            (18,	0.999008281)
            (20,	0.99343795)
            (22,	0.99343795)
            (24,	0.99343795)
            (26,	0.99343795)
            (28,	0.99343773)
            (30,	0.99343773)
            (32,	0.99343773)
            (34,	0.99343772)
            (36,	0.99343771)
            (38,	0.99343770)
            (40,	0.99343759)
            (42,	0.99343759)
            (44,	0.99343759)
            (46,	0.99343759)
            (48,	0.99343735)
            (50,	0.99343735)
            (52,	0.99343735)
            (54,	0.99343735)
            (56,	0.99343735)
            (58,	0.99343735)
            (60,	0.99343735)
            (62,	0.99343735)
            (64,	0.99343735)
            (66,	0.99343735)
            (68,	0.99343735)
            (70,	0.99343735)
            (72,	0.99343735)
            (74,	0.99343735)
            (76,	0.99343735)
            (78,	0.99343735)
            (80,0.648609459)
        	};
        	\addlegendentry{regularizer}
        	\end{axis}
        \end{tikzpicture}
    \end{minipage}
    \caption{Loss function value w.r.t. epochs. \textsc{MinCutPool} optimises the orthogonality loss, which decreases smoothly, while its min-cut objective remains constant (acting like a regularizer); whereas \textsc{HoscPool} optimises the main objective directly. Sometimes, \textsc{MinCutPool} does not manage to optimise the regularizer loss, yielding a degenerate clustering.}
    \label{fig:optim}
    \vspace{-.4em}
\end{figure}
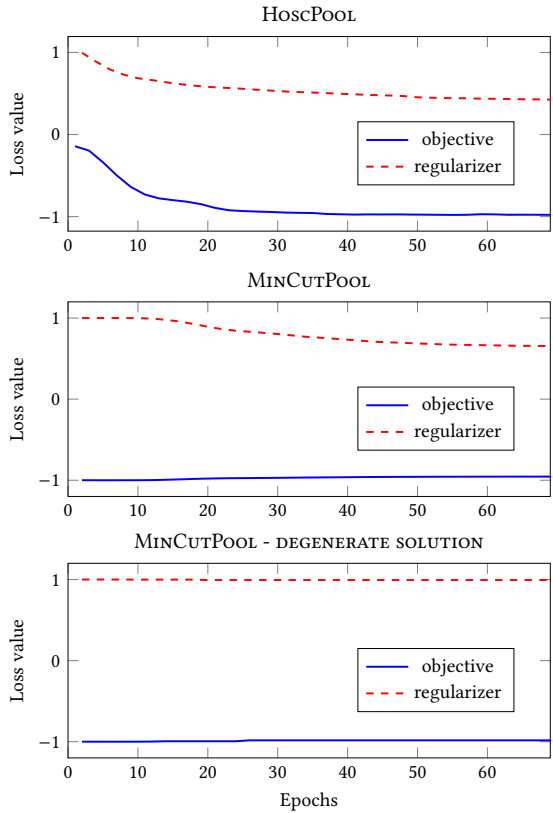

Before moving to the experiments, we take a moment to emphasise the key differences with respect to core end-to-end clustering-based pooling baselines. We focus on \textsc{MinCutPool} in the following since it is our closest baseline. \textsc{DiffPool} and others differ more significantly, in addition to being less theoretically-grounded and efficient.  

Firstly, \textsc{MinCutPool} focuses on first-order connectivity patterns, while we work on higher-order, which implies a more elaborated background theory with the construction and combination of several motif adjacency matrices (each specific to a particular motif). This shall lead to capturing more advanced types of communities, producing ultimately a better coarsening of the graph. Secondly, we approximate a probabilistic version of the motif conductance problem (extension of the normalised min-cut to motifs) whereas \textsc{MinCutPool} approximates the relaxed unormalised min-cut problem. Despite claiming to formulate a relaxation of the normalised min-cut (a trace ratio), it truly minimises a ratio of traces in the objective function: $- \frac{\text{Tr}(\S^\top \Tilde{\A}\S)}{\text{Tr}(\S^\top \Tilde{\D}\S)}$. Since $\text{Tr}(\S^\top\Tilde{\D}\S)=\sum_{i\in \mathcal{V}} \Tilde{D}_{ii}$ is a constant, this yields the unormalised min-cut $-\text{Tr}(\S^\top\Tilde{\A}\S)$, which often produces degenerate solutions. To cope with this limitation, \textsc{MinCutPool} optimises in parallel a penalty term $\mathcal{L}_o$ encouraging balanced and discrete clusters assignments. But despite this regularizer, it often gets stuck in local minima \cite{wang2007trace} (see Fig. \ref{fig:optim}), as we will see empirically in Section \ref{sec:eval}. We spot and correct this weakness in \textsc{HoscPool}. Thirdly, we introduced a new and more powerful orthogonality term together with a regularization control parameter. Unlike \textsc{MinCutPool}, it is unnecessary but often smoothes out training and improves performance. Lastly, we showcase a different architecture involving a more general way of computing $\S$. 

\section{Evaluation}
\label{sec:eval}
We now evaluate the benefits of the proposed method, with the goal of answering the following questions: 
\begin{enumerate}
    \item Does our differentiable higher-order clustering algorithm compute meaningful clusters? Is considering higher-order structures beneficial?
    \item How does \textsc{HoscPool} compare with state-of-the-art pooling approaches for graph classication tasks? 
    \item Why do existing pooling operators fail to outperform significantly random pooling? 
\end{enumerate}

\subsection{Clustering}
\label{subsec:clustering}

\begin{table*}[ht!]
\centering
\caption{(Right) NMI obtained by clustering the nodes of various networks over 10 different runs. Best results are in bold, second best underlined. The number of clusters $K$ is equal to the number of node classes. (Left) Dataset properties.} 
\label{tab:clust_cit}
\begin{tabular}{lcccc|ccccccc}
\cmidrule[1.3pt]{1-12}
\textbf{Dataset} & Nodes & Edges & Feat. & $K$ & \textsc{SC} & \textsc{MSC} & \textsc{DiffPool} & \textsc{MinCutPool} & \textsc{HP-1} & \textsc{HP-2} & \textsc{HoscPool} \\
\cmidrule[.5pt]{1-12}
\textsl{Cora}      & 2,708 & 5,429 & 1,433 & $7$ & $0.150$ \t{$\pm$ $0.002$} & $0.056$ \t{$\pm$ $0.014$} & $0.308$ \t{$\pm$ $0.023$} & $0.391$ \t{$\pm$ $0.028$} &  $0.435$ \t{$\pm$ $0.032$} & \underline{$0.464$} \t{$\pm$ $0.036$} & $\mathbf{0.502}$ \t{$\pm$ $0.029$}\\
\textsl{PubMed}  & 19,717 & 88,651 & 500     & $3$ & $0.183$ \t{$\pm$ $0.002$} & $0.002$ \t{$\pm$ $0.000$} & $0.098$ \t{$\pm$ $0.006$} & ${0.214}$ \t{$\pm$ $0.066$} &   \underline{$0.230$} \t{$\pm$ $0.071$} & ${0.215}$ \t{$\pm$ $0.073$} & $\mathbf{0.260}$ \t{$\pm$ $0.054$} \\ 
\textsl{Photo}  & 7,650 & 287,326 & 745    & $8$ & $\underline{0.592}$ \t{$\pm$ $0.008$} & $0.451$ \t{$\pm$ $0.011$} & $0.171$ \t{$\pm$ $0.004$} & $0.086$ \t{$\pm$ $0.014$} &  $0.495$ \t{$\pm$ $0.068$} & $0.513$ \t{$\pm$ $0.083$} & $\mathbf{0.598}$ \t{$\pm$ $0.101$}\\
\textsl{PC}   & 13,752 & 245,861 & 767    & $ 10 $ & $ 0.464 $ \t{$\pm$ $0.002$}  & $ 0.166$ \t{$\pm$ $0.009$}  & $0.043$ \t{$\pm$ $0.008$} & $ 0.026 $ \t{$\pm$ $0.006$}  &  $0.497$ \t{$\pm$ $0.040$} & \underline{${0.499}$} \t{$\pm$ $0.036$} & $\mathbf{0.528}$ \t{$\pm$ $0.041$} \\
\textsl{CS}  & 18,333 & 81,894 & 6,805      & $ 15 $ & $ 0.273 $ \t{$\pm$ $0.006$} & $ 0.011 $ \t{$\pm$ $0.009$} & $ 0.383 $ \t{$\pm$ $0.048$} & $ 0.431 $ \t{$\pm$ $0.060$} &  $0.479$ \t{$\pm$ $0.022$} & \underline{${0.701}$} \t{$\pm$ $0.029$} & $\mathbf{0.731}$ \t{$\pm$ $0.018$}\\
\textsl{Karate}    & 34 & 156 & 10   & $ 2 $ & $ 0.792 $ \t{$\pm$ $0.035$} & \underline{${0.870}$} \t{$\pm$ $0.031$} & $ 0.715 $ \t{$\pm$ $0.018$}  & $ 0.751 $ \t{$\pm$ $0.090$} &   $0.792$ \t{$\pm$ $0.038$} & ${0.862}$ \t{$\pm$ $0.046$} & $\mathbf{0.894}$ \t{$\pm$ $0.039$}\\
\textsl{DBLP}    & 17,716 & 105,734 & 1,639    & $ 4 $ & $ 0.027$ \t{$\pm$ $0.003$} & $ 0.005 $ \t{$\pm$ $0.006$} & $ 0.186 $ \t{$\pm$ $0.014$}  & $ \mathbf{0.334}$  \t{$\pm$ $0.026$}  &  \underline{$0.326$} \t{$\pm$ $0.027$} & $ 0.284 $ \t{$\pm$ $0.026$} & $0.312$ \t{$\pm$ $0.027$}\\
\textsl{Polblogs}     & 1,491 & 33,433 & 10  & $ 2 $ & $ 0.017 $ \t{$\pm$ $0.000$} & $ 0.014 $ \t{$\pm$ $0.001$} & $ 0.317 $ \t{$\pm$ $0.010$}  & $ 0.440 $ \t{$\pm$ $0.390$} &  \underline{$0.992$} \t{$\pm$ $0.003$} & $\mathbf{0.994} $ \t{$\pm$ $0.001$} & $\mathbf{0.994}$ \t{$\pm$ $0.005$}\\  
\textsl{Email-eu}     & 1,005 & 32,770 & 10  & $ 42 $ & $ 0.485 $ \t{$\pm$ $0.030$} & $ 0.382 $ \t{$\pm$ $0.019$} & $ 0.096 $ \t{$\pm$ $0.034$}  & $ 0.253 $ \t{$\pm$ $0.028$} &  $0.317$ \t{$\pm$ $0.026$} & $\mathbf{0.488}$ \t{$\pm$ $0.025$} & $\underline{0.476}$ \t{$\pm$ $0.021$}\\ \midrule
\textsl{Syn1}   & 1,000 & 6,243 & 10    & $ 3 $ & $ 0.000 $ \t{$\pm$ $0.000$} & $\mathbf{1.000} $ \t{$\pm$ $0.000$} & $ 0.035 $ \t{$\pm$ $0.000$}  & $ 0.043 $ \t{$\pm$ $0.008$} &  $ 0.041 $ \t{$\pm$ $0.006$} & $\mathbf{1.000}$ \t{$\pm$ $0.000$} & $\mathbf{1.000}$ \t{$\pm$ $0.000$}\\
\textsl{Syn2}    & 1,000 & 5,496 & 10   & $ 2 $ & $ 0.003 $ \t{$\pm$ $0.000$} & $ 0.050 $ \t{$\pm$ $0.003$} & $ 0.081 $ \t{$\pm$ $0.008$}  & $ 0.902 $ \t{$\pm$ $0.028$} &  $ 0.942 $ \t{$\pm$ $0.028$} & $\mathbf{1.000}$ \t{$\pm$ $0.000$} &  $\mathbf{1.000}$ \t{$\pm$ $0.000$} \\
\textsl{Syn3}    & 500 & 48,205 & 10    & $ 5 $ & $\mathbf{1.000} $ \t{$\pm$ $0.000$} & $\mathbf{1.000} $ \t{$\pm$ $0.000$} & $ 0.067 $ \t{$\pm$ $0.001$}  & $ 0.052 $ \t{$\pm$ $0.002$} & $ 0.115 $ \t{$\pm$ $0.006$}& \underline{${0.826}$} \t{$\pm$ $0.005$} & $\mathbf{1.000}$ \t{$\pm$ $0.000$} \\
\cmidrule[1.3pt]{1-12}
\end{tabular}
\vspace{-.5em}
\end{table*}

\textbf{Experimental setup}. For this experiment, we first run a Message Passing (MP) layer; in this case a \textsc{GCN} model with skip connection for initial features \cite{morris2019weisfeiler}: $\bar \X = \text{ReLU}(\A \X \mathbf{\Theta}_1 + \X \mathbf{\Theta}_2)$, where $\boldsymbol{\Theta}_1$ and $\boldsymbol{\Theta}_2$ are trainable weight matrices. It has $32$ hidden units and ReLU activation function. We then run a Multi-Layer Perceptron (\textsc{MLP}) with $32$ hidden units to produce the cluster assignment matrix of dimension $num\_nodes \times num\_clusters$, trained end-to-end by optimising the unsupervised loss function $\mathcal{L}_{mc} + \mu \mathcal{L}_{o}$. This architecture is trained using a learning rate of $0.001$ for an Adam optimizer, $500$ epochs, a gradient clip of $2.0$, $200$ early stop patience, a learning decay patience of $25$ and $\mu=\{0,.1, 1\}$. \\

\noindent \textbf{Metrics}. We evaluate the quality of $\S$ by comparing the distribution of true node labels with the one of predicted labels, via Normalised Mutual Information $\text{NMI}(\tilde{\y}, \y) = \frac{H(\tilde{\y}) - H(\tilde{\y}|\y)}{\sqrt{H(\tilde{\y}) - H(\y)}}$, where $H(\cdot)$ is the entropy and node cluster membership is determined by the argmax of its assignment probabilities. We also calculate completeness, modularity, normalised cut, and motif conductance (App. Table \ref{tab:clustering-all-metrics}). \\

\noindent \textbf{Datasets}. We use a collection of node classification datasets with ground truth community labels:  citation networks \textsl{Cora, PubMed}; collaboration networks \textsl{DBLP, Coauthor CS}; co-purchase networks \textsl{Amazon Photo, Amazon PC}; the \textsl{KarateClub} community network; and communication networks \textsl{Polblogs} and \textsl{Eu-email}. They are all taken from \href{https://pytorch-geometric.readthedocs.io/en/latest/modules/datasets.html}{Pytorch Geometric}. 
We construct three synthetic datasets: \textsl{Syn1}, \textsl{Syn2}, \textsl{Syn3} (based on several random graphs) where node labels are determined based on higher-order community structure and node features are simple graph statistics (Appendix \ref{sec:synthetic-data}). They are designed to show the additional efficiency of \textsc{HoscPool} when datasets have clear higher-order structure, which is not always the case for the standard baseline datasets chosen. \\


\noindent \textbf{Baselines.} We compare \textsc{HoscPool} with the original spectral clustering (SC), motif spectral clustering (MSC)\footnote{SC based on motif conductance \cite{benson2016higher} instead of edge conductance; meaning SC applied on $\A_M$.} as well as key pooling baselines \textsc{DiffPool} and \textsc{MinCutPool}. We refer to all methods by their pooling name for simplicity, although this experiment focuses on the clustering part and does not involve the coarsening step. We repeat all experiments 10 times and average results across runs. For ablation study, let \textsc{HP-1} and \textsc{HP-2} denote \textsc{HoscPool} where $\mathcal{L}_{mc}$ in Eq. (\ref{eq:mixedpool}) has $\alpha_2=0$ (first-order connectivity only) and $\alpha_1=0$ (higher-order only), respectively. \\

\textbf{Results} are reported in Table \ref{tab:clust_cit}. \textsc{HoscPool} achieves better performance than all baselines across most datasets. This trend is emphasised on synthetic datasets, where we know higher-order structure is critical, proving the benefits of our clustering method. \textsc{DiffPool} often fails to converge to a good solution. \textsc{MinCutPool}, as evoked earlier and in \cite{tsitsulin2020graph}, sometimes get stuck in degenerate solutions (e.g., \textsl{Amazon PC} and \textsl{Photo} -- all nodes are assigned to less than 10\% of clusters), failing completely to converge even when tuning model architecture and hyper-parameters (see Fig.\ref{fig:optim}). \textsc{HP-1} shows superior performance and alleviates this issue, meaning that it can be considered as an improved version of \textsc{MinCutPool}. Spectral Clustering (SC) performs really well on some datasets, poorly on others. \text{MSC} often performs badly, revealing its excessive dependence to the presence of motifs. On the contrary, our results highlight the robustness of \textsc{HoscPool} to the limited presence of motifs due to its consideration for node features. Besides, \textsc{HoscPool}'s consideration for finer granularity levels allows to group nodes primarily based on motifs while still considering edges when necessary, which may be the reason of its superior performance with respect to \textsc{HP-2}, itself more desirable than \textsc{HP-1} (edge-only). This ablation study proves the relevance of our underlying claims: incorporating higher-order information leads to better communities and combining several motifs further help. See Table \ref{tab:clustering-all-metrics} for more results.\\

\noindent \textbf{Complexity}. The main complexity of \textsc{HoscPool} lies in the derivation of $\A_M$, which remains relatively fast for triangles: $\A_M= \A^2 \odot \A$. In Table \ref{tab:time-clustering}, we remark that \textsc{HoscPool} (and \textsc{HP-2}) has a comparable running time with respect to \textsc{MinCutPool} on small or average size datasets. It is slower to compute than \textsc{MinCutPool} on large datasets, while staying relatively affordable. This extra time lies with the computation and processing of the motif adjacency matrix as well as the combination of several connectivity order; which grows bigger with the graph size. Note however that we could avoid the computation of the regularisation loss, which both \textsc{MinCutPool} and \textsc{DiffPool} cannot afford. \textsc{HP-1} is not reported as it shares similar times as \textsc{MinCutPool} while reaching better performance.

\begin{table}[h]
\caption{Running time (s) of the entire clustering experiment.} 
\label{tab:time-clustering}
\centering
\begin{tabular}{lcccc}
\cmidrule[1.3pt]{1-5}
\textbf{Dataset} & \textsc{DiffPool} & \textsc{MinCutPool} & \textsc{HP-2} & \textsc{HoscPool} \\
\cmidrule[.5pt]{1-5}
\textsl{Cora}      & $13$ & $16$ & $17$ & $24$ \\
\textsl{PubMed}  & $80$ & $95$ & $264$ & $501$\\
\textsl{Photo}      & $23$ & $48$ & $91$ & $182$\\
\textsl{PC}          & $89$ & $101$ & $304$ & $510$\\
\textsl{CS}          & $157$ & $251$ & $683$ & $1406$\\
\textsl{Karate}    & $9$ & $9$ & $9$ & $9$ \\
\textsl{DBLP}       & $126$ & $210$ & $635$ & $1330$\\
\textsl{Polblogs}      & $8$ & $9$ & $10$ & $10$\\
\textsl{Email-eu}      & $9$ & $9$ & $10$ & $12$\\ \midrule
%
%
\cmidrule[1.3pt]{1-5}
\end{tabular}
\end{table}


\begin{table*}[!t]
\centering
\caption{Graph classification accuracy. Top results are in bold, second best underlined.}
\vspace{-.6em}
\label{tab:graph_class}
\resizebox{\textwidth}{!}{
\begin{tabular}{lccccccccccc}

\cmidrule[1.3pt]{1-12}
\textbf{Dataset} & \textsc{NoPool} &  \textsc{Random} & \textsc{GMT} & \textsc{MinCutPool} & \textsc{DiffPool} & \textsc{EigPool} & \textsc{SAGPool} & \textsc{ASAP} & \textsc{HP-1} & \textsc{HP-2} &\textsc{HoscPool}  \\
\cmidrule[.5pt]{1-12}

\textsl{Proteins} & 71.6\t{$\pm$4.1} &  75.7\t{$\pm$3.2} & 75.0\t{$\pm$4.2} & 75.9\t{$\pm$2.4} & 73.8\t{$\pm$3.7} & 74.2\t{$\pm$3.1} & 70.6\t{$\pm$3.5} & 74.4\t{$\pm$2.6} & 76.7\t{$\pm$2.5} &  \underline{77.0}\t{$\pm$3.1} & \textbf{77.5}\t{$\pm$2.3} \\  
\textsl{NCI1} & 77.1\t{$\pm$1.9} & 77.0\t{$\pm$1.7} & 74.9\t{$\pm$4.3} & 76.8\t{$\pm$1.6} & 76.7\t{$\pm$2.1} & 75.0\t{$\pm$2.2} & 74.1\t{$\pm$3.9} & 74.3\t{$\pm$1.6} & 77.3\t{$\pm$1.6} & \textbf{80.3}\t{$\pm$2.0} & \underline{79.9}\t{$\pm$1.7}\\  
\textsl{Mutagen.}  & 78.1\t{$\pm$1.3} & 79.2\t{$\pm$1.3}  & 79.4\t{$\pm$2.2} & 78.6\t{$\pm$1.8} & 77.9\t{$\pm$2.3} & 75.2\t{$\pm$2.7} & 74.4\t{$\pm$2.7} & 76.8\t{$\pm$2.4} & 79.8\t{$\pm$1.6} & \underline{81.7}\t{$\pm$2.1} & \textbf{82.3}\t{$\pm$1.3} \\  
\textsl{DD} & 71.2\t{$\pm$2.2} & 77.1\t{$\pm$1.5} & 78.1\t{$\pm$3.2} & 78.4\t{$\pm$2.8} & 76.3\t{$\pm$2.1} & 75.1\t{$\pm$1.8} & 71.5\t{$\pm$4.1} & 73.2\t{$\pm$2.5} & \underline{78.8}\t{$\pm$2.0} & 78.2\t{$\pm$2.1} & \textbf{79.4}\t{$\pm$1.8} \\  
\textsl{Reddit-B} & 80.1\t{$\pm$2.6} & 89.3\t{$\pm$2.6} & 86.7\t{$\pm$2.6} & 89.0\t{$\pm$1.4} & 87.3\t{$\pm$2.4} & 82.8\t{$\pm$2.1} & 74.7\t{$\pm$4.5} & 84.1\t{$\pm$1.1}  & 91.2\t{$\pm$1.0} & \underline{92.8}\t{$\pm$1.5} & \textbf{93.6}\t{$\pm$0.9}\\  
\textsl{Cox2-MD} & 58.7\t{$\pm$3.2} & 62.9\t{$\pm$3.6} & 58.9\t{$\pm$3.6} & 58.9\t{$\pm$5.1} & 57.1\t{$\pm$4.8} & 59.8\t{$\pm$3.4} & 56.9\t{$\pm$9.7} & 60.5\t{$\pm$5.5} & 61.6\t{$\pm$3.5} & \textbf{66.4}\t{$\pm$4.6} & \underline{64.6}\t{$\pm$3.9} \\ 
\textsl{ER-MD} & 72.2\t{$\pm$2.9} & 73.0\t{$\pm$4.5} & 74.3\t{$\pm$4.5} & 75.5\t{$\pm$4.0} & 76.8\t{$\pm$4.8} & 73.1\t{$\pm$3.8} & 71.7\t{$\pm$8.2} & 74.5\t{$\pm$5.9} & 76.2\t{$\pm$4.2} & \underline{77.9}\t{$\pm$4.3} & \textbf{78.2}\t{$\pm$3.8} \\ 
\textsl{b-hard} & 66.5\t{$\pm$0.5} & 69.1\t{$\pm$2.1}  & 70.1\t{$\pm$3.4} & 72.6\t{$\pm$1.5} & 70.7\t{$\pm$2.0} & 69.1\t{$\pm$3.1} & 39.6\t{$\pm$9.6} & 70.5\t{$\pm$1.7} & 72.4\t{$\pm$0.8} & \underline{73.5}\t{$\pm$0.8} & \textbf{74.0}\t{$\pm$0.4} \\ 
\cmidrule[1.3pt]{1-12}
\end{tabular}
}
\end{table*}

\subsection{Supervised graph classification}
\label{subsec:graph-classification}

\textbf{Experimental setup}. We evaluate our pooling operator \textsc{HoscPool} on a plethora of graph classification (GC) tasks, for a fixed network architecture: GNN -- Pooling -- GNN -- Pooling -- GNN -- Global Pooling -- Dense ($\times$2). Again, the GNN chosen is a GCN with skip connection, as it was found more efficient than other GNNs (see ablation study in Table \ref{tab:ablation-study}). We sometimes add skip connections and global pooling to the output of the first and second GNN; and concatenate the resulting vector to the third GNN's output. Each MP layer and final dense layer has between $16$ and $64$ hidden units depending on the dataset regarded, and ReLU activation function. A Pooling block produces a cluster assignment matrix of dimension $num\_nodes \times \text{int}(num\_nodes * 0.25)$. The batch-size is different for every dataset, and ranges from $8$ to $64$. This architecture is trained using a learning rate for Adam of $0.001$, $500$ epochs, a gradient clip of $2.0$, $100$ early stop patience, a learning decay patience of $50$ and a regularisation parameter $\mu=\{0, 0.1\}$. \\

\noindent \textbf{Baselines.} We compare our method to representative state-of-the-art graph classification baselines, involving pooling operators \textsc{DiffPool} \cite{ying2018hierarchical}, \textsc{MinCutPool} \cite{bianchi2020spectral}, \textsc{EigPool} \cite{ma2019graph}, \textsc{SAGPool} \cite{lee2019self}, \textsc{ASAP} \cite{ranjan2020asap}, \textsc{GMT} \cite{baek2021accurate}; by replacing the pooling layer in the above pipeline. We implement a random pooling operator (\textsc{Random}) to assess the benefits of pooling similar nodes together, and a model with a single global pooling operator (\textsc{NoPool}) to assess how useful leveraging hierarchical information is. \\

\noindent \textbf{Datasets.} We use several common benchmark datasets for GC, taken from TUDataset \cite{morris2020tudataset}, including three bioinformatics protein datasets \textsl{Proteins}, \textsl{Enzymes}, \textsl{D$\&$D}; one mutagen \textsl{Mutagenicity}; one anticancer activity dataset \textsl{NCI1}; two chemical compound dataset \textsl{Cox-2-MD}, \textsl{ER-MD}; one social network \textsl{Reddit-Binary}. \textsl{Bench-hard} is taken from \href{https://github.com/FilippoMB/Benchmark_dataset_for_graph_classification}{source} where $\X$ and $\A$ are completely uninformative if considered alone. We split them into training set ($80\%$), validation set ($10\%$), and test set ($10\%$). We adopt the accuracy metric to measure performance and average the results over 10 runs, each with a different split. We select the best model using validation set accuracy, and report the corresponding test set accuracy. For featureless graphs, we use constant features. Model hyperparameters are tuned for each dataset, but are kept fixed across all baselines. Lastly, despite being used by all baselines, note that these datasets are known to be small and noisy, leading to large errors. \\


\textbf{Results} are reported in  Table \ref{tab:graph_class}, from which we draw the following conclusions. Performing pooling proves useful (\textsc{NoPool}) in most cases. \textsc{HoscPool} compares favourably on all datasets w.r.t. pooling baselines. Higher-order connectivity patterns are more desirable than first-order ones, and combining both is even better. It confirms findings from Section \ref{subsec:clustering} and shows that better clustering (i.e. graph coarsening) is correlated with better classification performance. However, while the clustering performance of \textsc{HoscPool} is significantly better than baselines, the performance gap has slightly closed down on this task. Even more surprising, the benefits of existing advanced node-grouping or node-dropping methods are not considerable with respect to the \textsc{Random} pooling baseline. Faithfully to what we announced in Section \ref{sec:intro}, we attempt to provide explanations. 



\subsection{Pooling behaviour investigated}
\label{subsec:pooling-investigated}

First of all, we investigate the optimisation process of some key pooling operators (e.g., \textsc{MinCutPool}, \textsc{DiffPool}). We notice that they do not really learn to optimise their cluster assignment matrix on these graph classification tasks, producing degenerate solutions where most nodes are assigned to few clusters (similarly to Fig.\ref{fig:optim}). This issue would explain why random pooling performs on par with them; as they do not learn structurally meaningful clusters. 

A potential solution to this problem is to design a clustering-based pooling operator allowing to capture faithfully a more advanced kind of relationship between nodes, which we tried to do with \textsc{HoscPool}. We also tested a variety of architectures and optimisation options to see if learning would occur in specific situations. For instance, we tested several GNNs, different model architectures, skip-connections, no supervised loss at the start, etc. (see ablation study in Table \ref{tab:ablation-study}). However, despite clear progress -- we learn to decently optimise $\S$, to assign nodes to more clusters and to better balance the number of nodes per cluster -- there still seems to be room for improvement. We thus look for other potential causes which could prevent a proper learning, especially targeting the graph classification model architecture and the nature of selected datasets.

Concerning model architecture, we show in Appendix \ref{subsec:two-layers-clustering} that using more complex clustering frameworks (\textit{2-layer clustering}: GNN -- Pooling -- GNN --Pooling) prevents totally the learning of meaningful clusters for \textsc{MinCutPool} (and \textsc{DiffPool}), which illustrates a feature oversmoothing issue. \textsc{HoscPool}, on the other hand, has fixed this issue and still manages to learn meaningful clusters. Nevertheless, the learning process becomes longer and more difficult, leading to a drop in performance. In addition to showing the robustness of \textsc{HoscPool} with respect to existing pooling baselines, this experiment reveals that the clustering performed in graph classification tasks may not lead to meaningful clusters because of the more complex framework. Although it is likely to contribute, it is probably a factor among others, since simpler GC models like GNN -- Pooling -- GNN -- Global Pooling -- Dense (\textit{1-pooling} in Table \ref{tab:ablation-study}) do not improve things. 

We therefore also look for answers from a dataset perspective. In Table \ref{tab:GC_datasets_prop}, the computed graph properties and clustering results on individual graphs suggest that graphs are relatively small, with few node types co-existing in a same graph, weak homophily and a relatively poor community structure which clustering algorithms would like to exploit. Besides, because most datasets do not have dense node features (only labels), the node identifiablity assumption is shaken and does not enable our MLP (\ref{eq:assigments}) to fully distinguish between same-label-nodes, thus making it impossible to place them in distinct clusters. On top of that, we now need to learn a clustering pattern that extends to all graphs, which is a much more complex task (compared to 1 graph in Section \ref{subsec:clustering}).

\begin{table*}[h]
    \caption{(Left) Simple graph statistics. (Middle) The clustering coefficient (\textit{cc}), proportion of triangles attached per node (\textit{triangle}), transivitiy (\textit{transi}), homophily (\textit{homo}) and proportion of node labels in a graph w.r.t. all graphs (\textit{diff labels}) are computed on each graph individually and averaged over the whole dataset. (Right) \textit{msc, sc} and \textit{sc-mod} denote motif conductance, normalised cut, and modularity obtained by clustering each graph using traditional deterministic spectral clustering, where the number of clusters is equal to the number of labels in a graph. The last column refers the NMI obtained through \textsc{HoscPool} clustering only. All metrics provide information on graph community structure. \textsl{Reddit-Binary} has no node labels and is treated differently. 
    }
    \vspace{-.6em}
    \label{tab:GC_datasets_prop}
    \centering
    \begin{tabular}{lcccc|ccccc|cccc}
    \midrule[1.3pt]
         Datasets & \# graphs & \# edges & av \# nodes & labels & \textit{cc} & \textit{triangle} & \textit{transi} & \textit{homo} & \textit{diff-labels} & \textit{msc} & \textit{sc} & \textit{sc-mod} & \textit{NMI}  \\
         \midrule
         Proteins  & 1,113 & 162,088 & 39 & 3 & .575 & 1.03 & .517 & .476 & .833 & .034 & .005 & .460 & .46 \\
         NCI1  & 4,110 & 132,753 & 29  & 37 & .125 & .125 & .214 & .667 & .054 & .111 & 0.0 & .388 & .71\\
         DD  & 1,178 & 843,046 & 284 & 89 & .496 & 2.0 & .462 & .058 & .219 & .021 & .013 & .402 & .38 \\
         Mutagenicity & 4,337 & 133,447 & 30 & 14 & .002 & .003 & .002 & .376 & .244 & .056 & 0.0 & .378 & .85 \\
         Reddit-Binary & 2,000 & 995,508 & 429 & no & .051 & .069 & .009 & - & - & .008 & .011 & .071 & -\\ %
         COX2-MD & 303 & 203,084 & 26.2 & 7 & 1.00 & 103 & 1.00 & .707 & .482 & .302 & .333 & .01 & .45 \\
         ER-MD & 446 & 209482 & 21.1 & 10 & 1.00 & 77.4 & 1.00 & .701 & .232 & .331 & .323 & .01 & .56 \\
    \midrule[1.3pt]
    \end{tabular}
\end{table*}

\begin{table*}[!t]
\caption{Ablation study of \textsc{HoscPool}, denoted as \textsl{Base}. \textsl{GIN}, \textsl{SAGE}, \textsl{GAT} change the core GNN model; \textsl{No-diag} does not zero-out the diagonal of $\S$ in the pooling step, \textsl{1-pooling} uses an architecture with only one \textsc{HoscPool} block, \textsl{skip-co} adds a skip connection between every GNN layer and the dense layer, \textsl{c-ratio} involves a higher clustering ratio and \textsl{no-adapt} refers to the discussed dynamic adaptative loss. For \textsl{dense-feat}, we simply added some graph statistics to boost node identifiability. }
\vspace{-.6em}
\label{tab:ablation-study}
\centering
\resizebox{\textwidth}{!}{
\begin{tabular}{p{0.08\textwidth} P{0.08\textwidth}P{0.08\textwidth}P{0.08\textwidth}P{0.08\textwidth}P{0.09\textwidth}P{0.09\textwidth} P{0.08\textwidth} P{0.08\textwidth}}
\cmidrule[1.3pt]{1-9}
\textbf{Model} & \textsc{Proteins} & \textsc{NCI1} & \textsc{Mutagen.} & \textsc{DD} & \textsc{\small{Reddit-B}} & \textsc{\small{Cox2-MD}} & \textsc{ER-MD} & \textsc{b-hard} \\
\cmidrule[.5pt]{1-9}
\textsl{Base} & \textbf{77.5} & \textbf{79.9} & \textbf{82.3} & \textbf{79.4} & \textbf{93.6} & \textbf{64.6} & \textbf{78.2} & \textbf{74.0} \\  
\textsl{$\mu=0$} & 76.4 & 78.9 & 80.7 & 78.1 & 93.6 & 64.2 & 75.4 & 73.1 \\  
\textsl{not-ada} & 77.5 & 78.9 & 80.8 & 79.4 & 93.4 & 62.4 & 76.4 & 72.5 \\  
\textsl{No-diag} & 77.7 & 77.2 & 80.0 & 78.9 & 90.2 & 60.9 & 74.6 & 70.7 \\  
\textsl{SAGE} & 76.7 & 77.2 & 79.5    & 78.9 & 92.4 & 62.1 & 74.9 & 71.0 \\  
\textsl{GAT} & 77.6 & 78.6 & 77.9     & 79.2 & 91.5 & 60.6 & 73.4 & 74.4 \\  
\textsl{GIN} & 76.9 & 77.7 & 76.7     & 79.6 & 93.6 & 58.7 & 77.0 & 71.5 \\  
\textsl{skip-co} & 77.2 & 77.7 & 80.5 & 79.5 & 93.9 & 61.8 & 76.6 & 71.8 \\  
\textsl{1-pooling} & 76.6 & 79.9 & 82.3& 78.3 & 90.5 & 63.6 & 77.4 & 74.0 \\   
\textsl{c-ratio} & 75.1 & 77.4 & 80.3 & 78.9 & 92.3 & 61.6 & 75.3 & 70.4 \\    
\textsl{dense-feat}& 77.2 & 79.4 & 80.0& 78.7 & 92.0 & 58.5 & 73.2 & 70.8 \\  
\cmidrule[1.3pt]{1-9}
\end{tabular}
}
\end{table*}

As a result, taking into consideration the multiple pooling layers, the joint optimisation with a supervised loss, the poor individual graph community structure, and the complexity of learning to cluster all graphs with few features, learning meaningful clusters becomes extremely challenging. This would explain the optimisation difficulties encountered by existing pooling operators so far. Although \textsc{HoscPool} makes a step towards better pooling, we advice future research to explore more appropriate datasets than TUDataset \cite{morris2020tudataset} even though it is used by all pooling baselines as benchmark, such as Open Graph Benchmark datasets (\href{https://ogb.stanford.edu/}{OGB}). We also recommend to design simpler node-grouping approach, to use higher-order information so as to capture more relevant communities even with complex model architectures, as well as to exploit more directly graph structure information (as targeted graphs do not have dense node features). Finally, the heterophilious nature of these datasets (Table \ref{tab:GC_datasets_prop}) come to question the true benefit of grouping together nodes with similar embeddings (homophily assumption) when coarsening the graph.

\section{Conclusion}

We have introduced \textsc{HoscPool}, a new hierarchical pooling operator bringing higher-order information to the graph coarsening step, ultimately leading to motif-aware hierarchical graph representations. \textsc{HoscPool} builds on a novel end-to-end clustering scheme, which designs an objective function combining several continuous relaxations of motif spectral clustering, avoiding the shortcomings of deterministic methods and solving the limitations of previous key baselines \textsc{DiffPool} and \textsc{MinCutPool}. The proposed experiments, through cluster observation and pooling performance, demonstrate the advantages brought by considering higher-order connectivity patterns and by combining flexibly different levels of motifs. Finally, our discussion about the relevance of the pooling operation itself aims to inspire and guide future research to design more adapted and efficient pooling operators, ensuring significant improvement over the random baseline for graph classification tasks. \\

\noindent \textbf{Acknowledgements.} Supported in part by ANR (French National Research Agency) under the JCJC project GraphIA (ANR-20-CE23-0009-01).



\newpage

\appendix
\section{Synthetic datasets}
\label{sec:synthetic-data}
(1) \textsl{syn1} is made of $k$ communities, each densely intra-connected by triangles. We then widely link these communities without creating new triangles through these new links. We create random Gaussian features (included one correlated to node labels) since our method is dependent on node features.

\noindent (2) \textsl{syn2} is an Erdős–Rényi random graph with 1,000 nodes and $p=0.012$. Each node receives label 0 if it does not belong to a triangle and label 1 otherwise. Node features include several graph statistics. 

\noindent  (3) \textsl{syn3} is designed using a Gaussian random partition graph with $k$ partitions with size drawn from a normal distribution. Nodes within the same partition are connected with probability $p=0.8$, while nodes across partitions with probability $0.2$. Here, only random features are used.


\section{2-layer clustering: precisions}
\label{subsec:two-layers-clustering}

In this experiment, we complexify the clustering framework (MP -- MLP), making it more similar to its use as a pooling operator inside supervised graph classification tasks. More precisely, we follow an architecture: MP -- Pooling -- MP -- Pooling. As before, the pooling step regroups an MLP to compute the first cluster assignment matrix $\S_1$, and a graph coarsening step. In the end, we provide a unique cluster assignment matrix $\S$ of dimension $N \times K$, composed of the two matrix derived above ($\S_1$ and $\S_2$), such that the probability that node $i$ belongs to cluster $k$ is written $\S_{ik} = \sum_j \S_{1_{ij}} \S_{2_{jk}}$.

The results, given in Table \ref{tab:2-layer-clustering}, are obtained using $1,000$ epochs with $early\_stop\_patience=500$---meaning using many more epochs than for standard 1-layer clustering. This is because the convergence to a desirable solution is weaker. Furthermore, the obtained solution is less desirable and yields to a less desirable clustering. Overall, this argument is very important as it suggests that the clustering obtained in supervised graph classification tasks might not be as accurate as what our original evaluation on real-world dataset with ground-truth community structure suggested. 

\begin{table}[h!]
\centering
\begin{tabular}{lccc}
\cmidrule[1.3pt]{1-3}
\textbf{Dataset} & \textsc{MinCutPool} & \textsc{HoscPool} \\
\cmidrule[.5pt]{1-3}
\textsl{Cora}     & $0.000$ \t{$\pm$ $0.000$} & $0.369$ \t{$\pm$ $0.026$}  \\
\textsl{PubMed}      & $0.000$ \t{$\pm$ $0.000$} & ${0.187}$ \t{$\pm$ $0.013$} \\ 
\textsl{Photo}      & $0.007$ \t{$\pm$ $0.005$} & $0.230$ \t{$\pm$ $0.063$} \\
\textsl{PC}  & $0.004$ \t{$\pm$ $0.001$}  & ${0.194}$ \t{$\pm$ $0.042$}  \\
\textsl{CS}       & $ 0.446$ \t{$\pm$ $0.018$} & ${0.417}$ \t{$\pm$ $0.025$} \\
\textsl{Karate} & $ 0.000$ \t{$\pm$ $0.000$}  &  $0.745$ \t{$\pm$ $0.046$} \\
\textsl{DBLP}    & $0.174$  \t{$\pm$ $0.078$}  & $ 0.244 $ \t{$\pm$ $0.023$} \\
\textsl{Polblogs}  & $0.135$ \t{$\pm$ $0.101$} & ${0.994} $ \t{$\pm$ $0.003$} \\  
\textsl{Email-eu}    & $0.197$ \t{$\pm$ $0.016$} & ${0.421}$ \t{$\pm$ $0.009$} \\ \midrule
\cmidrule[1.3pt]{1-3}
\end{tabular}
\caption{NMI of \textsc{MinCutPool} and \textsc{HoscPool} for 2-layer clustering framework}
\label{tab:2-layer-clustering}
\vspace{-.5em}
\end{table}

\begin{table*}[h]
\caption{Modularity (Mod), Conductance (Cond), Motif Conductance (M.Cond), Homogeneity (Homog) obtained by clustering the nodes of various networks over 10 different runs. The number of clusters $K$ is equal to the number of node classes. \textsc{HP-2} optimises better the motif conductance metric than \textsc{MinCutPool}. \textsc{HoscPool} achieves a similar motif conductance but a better conductance than  \textsc{HP-2}, which it also often outperforms in terms of modularity. Finally, \textsc{MinCutPool} does achieve degenerate solutions for several datasets (e.g., \textsl{PC, Photo, CS, Email-eu}). } 
\label{tab:clustering-all-metrics}
\footnotesize
\centering
\resizebox{\textwidth}{!}{
\begin{tabular}{lcccc|cccc|cccc}
\cmidrule[1.3pt]{1-13}
& \multicolumn{4}{c}{\scriptsize{\textsc{MinCutPool}}} &  \multicolumn{4}{c}{\scriptsize{\textsc{HP-2}}} &  \multicolumn{4}{c}{\scriptsize{\textsc{HoscPool}}}\\
    \cmidrule(l){2-5} \cmidrule(l){6-9} \cmidrule(l){9-13} 
\textbf{Dataset} & Mod & Cond & M.Cond & Homog & Mod & Cond & M.Cond & Homog & Mod & Cond & M.Cond & Homog \\
\cmidrule[.5pt]{1-13}
\textsl{Cora}   & $0.700$ & $0.156$ & $0.094$ & $0.464$ & $0.621$ & $0.125$ & $0.025$ & $0.338$ & $0.654$ & $0.091$ & $0.026$ & $0.314$ \\
\textsl{PubMed}   & $0.532$ & $0.120$ & $0.047$ & $0.225$ & $0.478$ & $0.069$ & $0.029$ & $0.101$ & $0.454$ & $0.082$ & $0.038$ & $0.096$ \\
\textsl{CS}   & $-0.005$ & $0.001$ & $0.000$ & $0.000$ & $0.684$ & $0.141$ & $0.087$ & $0.637$ & $0.695$ & $0.131$ & $0.084$ & $0.638$ \\
\textsl{Photo}  & $0.000$ & $0.008$ & $0.002$ & $0.002$ & $0.566$ & $0.084$ & $0.033$ & $0.470$ & $0.684$ & $0.093$ & $0.043$ & $0.580$ \\
\textsl{PC}  & $-0.001$ & $0.000$ & $0.000$ & $0.000$ & $0.546$ & $0.285$ & $0.263$ & $0.457$ & $0.591$ & $0.149$ & $0.082$ & $0.556$ \\
\textsl{DBLP}  & $0.533$ & $0.182$ & $0.157$ & $0.363$ & $0.588$ & $0.131$ & $0.065$ & $0.277$ & $0.608$ & $0.114$ & $0.066$ & $0.318$ \\
\textsl{Karate}   & $0.370$ & $0.269$ & $0.281$ & $0.543$ & $0.389$ & $0.192$ & $0.088$ & $0.715$ & $0.417$ & $0.217$ & $0.133$ & $0.861$ \\
\textsl{Email-eu}   & $0.002$ & $0.011$ & $0.003$ & $0.025$ & $0.189$ & $0.455$ & $0.382$ & $0.166$ & $0.185$ & $0.488$ & $0.396$ & $0.208$ \\
\textsl{Polblogs}   & $0.409$ & $0.090$ & $0.048$ & $0.991$ & $0.409$ & $0.087$ & $0.035$ & $0.993$ & $0.429$ & $0.073$ & $0.035$ & $0.991$ \\
\cmidrule[1.3pt]{1-13}
\end{tabular}
}
\end{table*}


\section{Extension to 4-nodes motifs}
\label{subsec: extension-4-nodes}
Here, we consider motifs composed of 4 nodes ($|\mathcal{M}|=4$), such as the 4-cycle or $K_4$, written as $\mathbf{v} = \{l,q,r,k\}$. In Section \ref{subsec:motif-spectral-clustering}, we formulated a relation between triangle normalised cut and graph-normalised cut, in order to compute triangle normalised cut easily. Here, we do the same, but for 4-nodes-motif conductance. Again, we derive this relation by looking at a single cluster $\mathcal{S}$ with corresponding cluster assignment vector $\mathbf{y}$, with $y_i=\begin{cases}
1 \text{ if } i \in \mathcal{S} \\
0 \text{ else}
\end{cases}$.

\begin{align*}
    3 \texttt{cut}^{(\mathcal{G})}_M(\mathcal{S}, \bar{\mathcal{S}}) &= 3 \sum_{\mathbf{v} \in \mathcal{M}} \mathbf{1}\{\exists i,j \in \mathbf{v} | i \in \mathcal{S}, j \in \bar{\mathcal{S}}\} \\
    =&  \sum_{\mathbf{v} \in \mathcal{M}} \big[ 3(y_l+y_q+y_r+y_k) \\
    &- 2(y_ly_q + y_ly_r + y_ly_k +y_qy_r+ y_qy_k+y_ry_k) \\
    &- \mathbf{1}\{\text{exactly 2 of } l,q,r,k \text{ are in $\mathcal{S}$}\}\big]. 
\end{align*}
This expression equals $\sum_{\mathbf{v} \in \mathcal{M}}
\begin{cases}
0 \text{ if all } y_l, y_q, y_r, y_k \text{ are the same} \\
3 \text{ if 3 are the same} \\
4 \text{ if 2 are the same} .
\end{cases}$

\noindent Thus,
\begin{align*}
    3 \texttt{cut}^{(\mathcal{G})}_M(\mathcal{S}, \bar{\mathcal{S}}) &+  \mathbf{1}\{\text{exactly 2 of $l,q,r,k$ are in $\mathcal{S}$}\} \\ 
    &= \sum_{\mathbf{v} \in \mathcal{M}} \big[ 3 (y_l + y_q + y_r + y_k) \\
    &- 2 (y_ly_q + y_ly_r + y_ly_k +y_qy_r+ y_qy_k+y_ry_k)\big] \\
    &= \mathbf{y}^\top \D_M\mathbf{y} - \mathbf{y}^\top \A_M\mathbf{y} \\
    &= \mathbf{y}^\top \L_M\mathbf{y} \\
    &= \texttt{cut}^{(\mathcal{G}_M)}(\mathcal{S}, \bar{\mathcal{S}}),
\end{align*}
where the second inequality holds because 
\begin{align*}
    \mathbf{y}^\top  \D_M\mathbf{y} &= \sum_{i \in \mathcal{S}} \sum_{j \in \mathcal{V}}  (A_M)_{ij}  = \texttt{vol}^{(\mathcal{G}_M)}(\mathcal{S}) = |\mathcal{M}| \texttt{vol}^{(\mathcal{G})}_M(\mathcal{S}) \\
    &=  3 \texttt{vol}^{(\mathcal{G})}_M(\mathcal{S}) = 3 \sum_{\mathbf{v} \in \mathcal{M}} (y_l + y_q + y_r + y_k)
\end{align*}
\begin{align*}
    \mathbf{y}^\top  \A_M\mathbf{y} &= \sum_{i \in \mathcal{V}} y_i \sum_{j \in \mathcal{V}}  y_j (A_M)_{ij}  = \sum_{i,j \in \mathcal{S}} (A_M)_{ij} =  \sum_{i,j \in \mathcal{S}} \sum_{\mathbf{v} \in \mathcal{M}} \mathbf{1}\{i,j \in \mathbf{v}\} \\
    &= \sum_{\mathbf{v} \in \mathcal{M}} 2 (y_ly_q + y_ly_r + y_ly_k +y_qy_r+ y_qy_k+y_ry_k).
\end{align*}

Overall, we obtain the following equality: 
\begin{align*}
     \texttt{cut}^{(\mathcal{G})}_M(\mathcal{S}, \bar{\mathcal{S}}) = \frac{1}{3} \texttt{cut}^{(\mathcal{G}_M)}(\mathcal{S}, \bar{\mathcal{S}}) - \frac{1}{3} \sum_{\mathbf{v} \in \mathcal{M}} \mathbf{1}\{\text{exactly 2 of $l,q,r,k$} \in \mathcal{\mathcal{S}}\}
\end{align*}

The optimisation problem can be written as: 
\begin{align*}
    &\min_\mathcal{S} \sum_k \frac{\texttt{cut}_M^{(\mathcal{G})}(\mathcal{S}_k, \bar{\mathcal{S}_k})}{\texttt{vol}_M^{(\mathcal{G})}(\mathcal{S}_k)} \\
    \equiv& \min_\mathcal{S} \sum_k \frac{\frac{1}{3} \texttt{cut}^{(\mathcal{G}_M)}(\mathcal{S}_k, \bar{\mathcal{S}_k}) - \frac{1}{3} \sum_{\mathbf{v} \in \mathcal{M}} \mathbf{1}\{\text{exactly 2 nodes in \textbf{v}  $\in \mathcal{S}$}\}}{\frac{1}{3}\texttt{vol}^{(\mathcal{G}_M)}(\mathcal{S}_k)} \\
    \equiv& \min_{\S \in [0,1]^{N\times K}} -\text{Tr}\bigg(\frac{\S^\top \A_M\S}{\S^\top \D_M\S}\bigg) - \sum_k \frac{ \sum_{\mathbf{v} \in \mathcal{M}} \mathbf{1}\{\text{exactly 2 nodes in \textbf{v} $\in \mathcal{S}$}\} }{\texttt{vol}^{(\mathcal{G}_M)}(\mathcal{S}_k)}.
\end{align*}

In practice however, unlike triangle normalised cut, this expression is not easy to compute. First of all, computing the related motif adjacency matrix is difficult; it cannot be written a simple matrix dot product. Secondly, there is this term on the RHS to take into consideration. And although we might be able to compute both directly via a complex algorithm, it is not guaranteed that solving this problem is quicker than the original optimisation problem (def. of $\texttt{vol}^{(G)}_M$ and  $\texttt{cut}^{(G)}_M$).\\

\bibliographystyle{ACM-Reference-Format}
\bibliography{main}

\end{document}